\documentclass[10pt,twocolumn,letterpaper]{article}

\usepackage{iccv}
\usepackage{times}
\usepackage{epsfig}
\usepackage{graphicx}
\usepackage{amsmath}
\usepackage{amssymb}
\usepackage{booktabs}
\usepackage{verbatim}

\usepackage[breaklinks=true,bookmarks=false]{hyperref}

\iccvfinalcopy 

\def\iccvPaperID{****} 

\begin{document}

\title{A Technical Report for ICCV 2021 VIPriors Re-identification Challenge}

\author{Cen Liu, Yunbo Peng, Yue Lin\\
NetEase Games AI Lab\\
{\{liucen01, gzpengyunbo, gzlinyue\}}@corp.netease.com\\
}

\maketitle

\begin{abstract}
   Person re-identification has always been a hot and challenging task. This paper introduces our solution for the re-identification track in VIPriors Challenge 2021. In this challenge, the difficulty is how to train the model from scratch without any pretrained weight. In our method, we show use state-of-the-art data processing strategies, model designs, and post-processing ensemble methods, it is possible to overcome the difficulty of data shortage and obtain competitive results. (1) Both image augmentation strategy and novel pre-processing method for occluded images can help the model learn more discriminative features. (2) Several strong backbones and multiple loss functions are used to learn more representative features. (3) Post-processing techniques including re-ranking, automatic query expansion, ensemble learning, etc., significantly improve the final performance. The final score of our team (ALONG) is 96.5154\% mAP, ranking first in the leaderboard.
\end{abstract}

\section{Introduction}
2021 VIPriors Re-identification Challenge is one of ``2nd Visual Inductive Priors for Data-Efficient Deep Learning Workshop" in the ICCV2021 conference. It focuses on obtaining high mean Average Precision (mAP) on a dataset coming from short sequences of basketball games. This dataset contains 8,569 person images of 436 identities in the training set while testing data is composed of the 9,171 images of 468 identities.

Person re-identification (ReID) aims at recognizing pedestrians across non-overlapping camera views, which draws wide attention due to its wide applications in surveillance, tracking, smart retail, etc. ~\cite{zheng2015scalable, zhao2017deeply, sun2018beyond, liu2021hlfnet}. As deep learning prevails, the CNN based ReID methods progress rapidly and achieve impressive performance on benchmark datasets. However, this challenge restricts the use of external data and pre-trained weights, due to the small dataset, it is difficult to perform well with general training due to the small dataset.. 

We solve this problem by experimenting with various techniques. In our method, based on strong backbones, we use multiple loss functions, data augmentation strategies, and ensemble learning to improve the performance.

\begin{figure}[t]
\centering 
\includegraphics[width=0.45\textwidth]{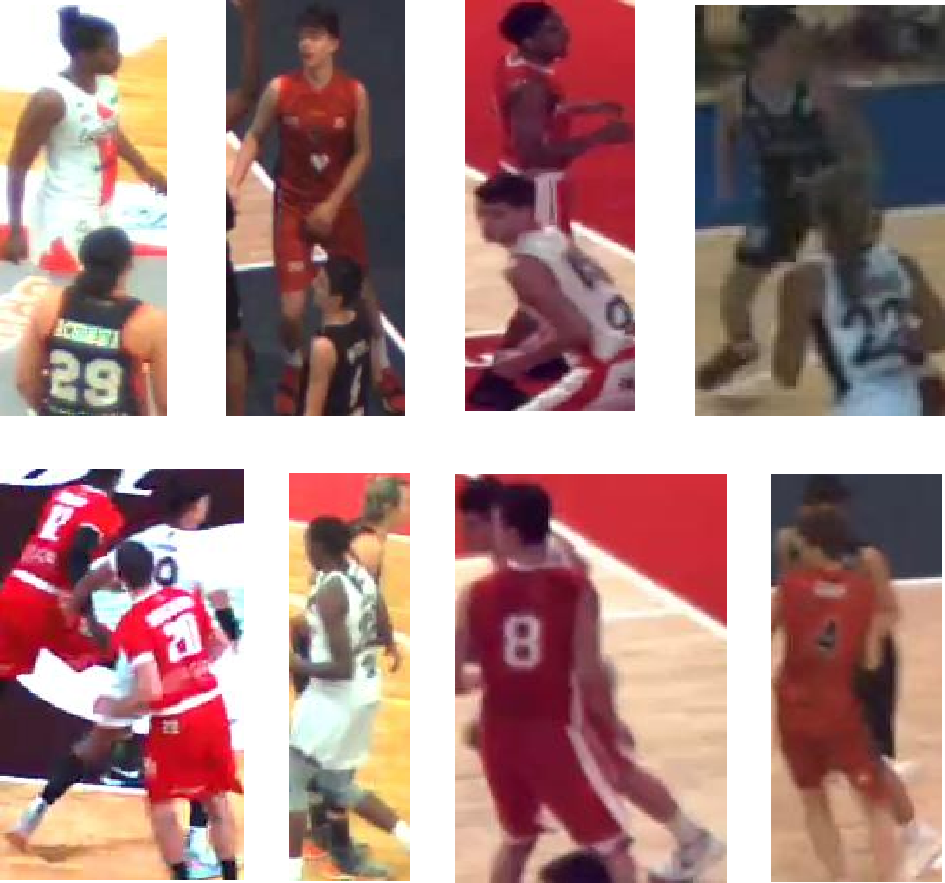}
\caption{Illustration of same noise data samples with varying degrees of occlusion. The is partial occlusion is in the first row; the second row is completely occluded (the label of the images in the second row is the person occluded).}
\label{fig:fig1}
\end{figure}

\begin{figure*}[t]
\centering 
\includegraphics[width=0.98\textwidth]{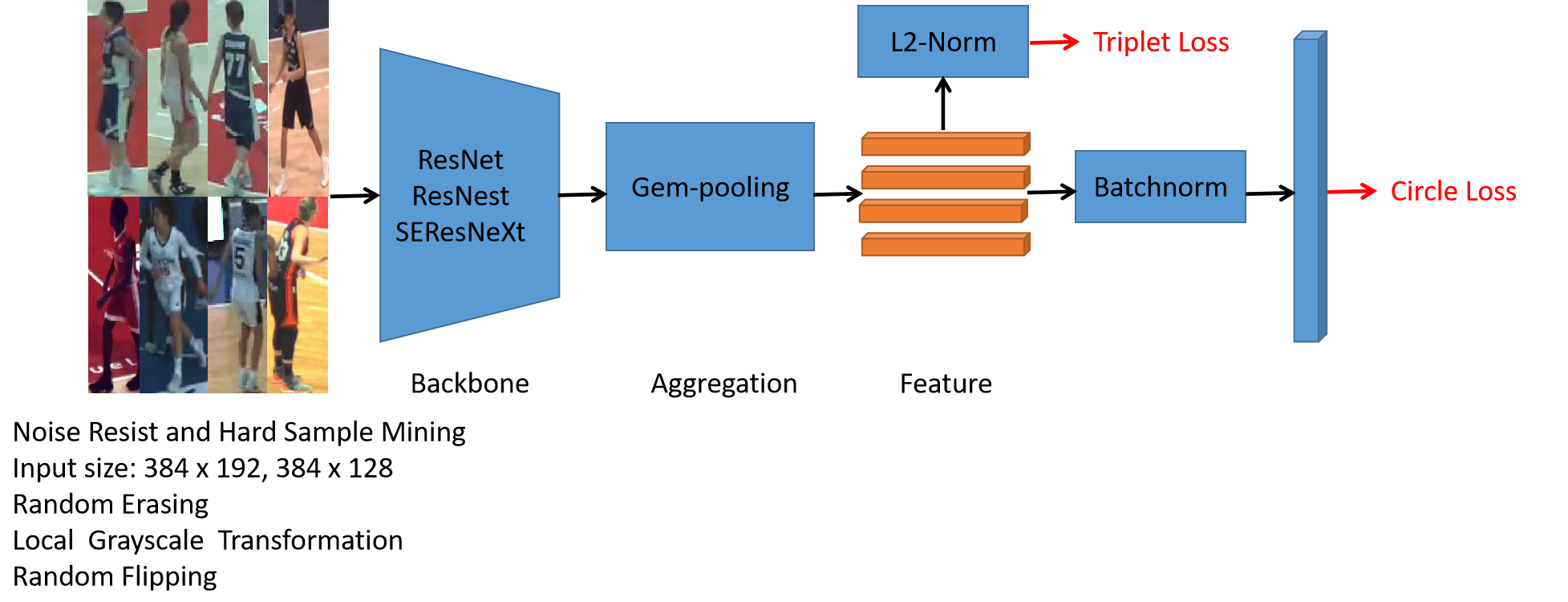}
\caption{Illustration of our proposed framework.}
\label{fig:fig2}
\end{figure*}

First, we focus on the data augmentation in data pre-processinging such as Random Erasing~\cite{zhong2020random}, Local Grayscale Transformation~\cite{Gong2021AGM} and random flip horizontally, etc., to improve the diversity of person images. In addition, there are different degrees of occlusion in the raw data, which lead to the problem of noise in network training and result in hard samples. To address these challenges, we use hard sampling method to select difficult samples and clean noise samples. Then, we apply strong backbones, multiple loss functions and add some training tricks by this paper~\cite{luo2019bag} to enable model training well with even a small amount of data. Finally, we apply automatic query expansion (AQE)~\cite{chum2007total} and re-rank~\cite{zhong2017re}, and post-processing ensemble techniques, etc. to compensate for the difficult problems through various models and multi-scale testing to produce good performance. Through the model applies with these techniques, our approach can significantly improve the ReID performance and achieve a competitive result on the test set.  According to the rules of the competition, we do not use any external image/video data or pre-trained weights. The implementation details of our solution are described in section 2 and section 3.

\section{Our Method}
In this section, our method for this challenge is introduced in detail. The critical parts of our method include data pre-processing, model architecture, loss function and post-processing techniques, which are elaborated in detail as follows.

\subsection{Pre-processing}
Due to the limited quantity and quality of original data, effective data pre-processing technology can improve the recognition performance of the model.

\textbf{Noise Resist and Hard Sample Mining:} Most of the person re-identification datasets contain a small portion of noisy labels, which requires the learning algorithm to resist certain amount of noises. Through our observation of this dataset, we find that there is also a lot of labeling noise. Therefore, we choose the online difficult sample mining algorithm proposed by Shrivastava et al.~\cite{shrivastava2016training} to mine the ``hard sample (noise)" data according to the loss of the training set in the training stage. The noise data we filtered out is shown in Figure \ref{fig:fig1}. We can see that this part of the data has different degrees of person occlusion.

Our initial approach is to delete this part of ``noise" data, but we find that if all occlusion data are deleted through experiments, the results of the test set fall rather than rise. We find the main reason is that there is occlusion data in the test set, but the occlusion is not serious (not complete occlusion). As shown in figure \ref{fig:fig1}, the partial occlusion is in the first row; the second row is completely occluded (the label of the images in the second row is occluded by person). Therefore, inspired by ~\cite{shrivastava2016training}, we divide samples with different degrees of occlusion into partial occlusion and full occlusion, according to the training loss and the selected threshold. Our approach is to delete completely occluded samples as noise data. The data with slight occlusion is not deleted, but as the ``hard samples" in training stage, and several data augmentations are used for the “hard samples”, increasing the proportion of hard samples in the original training set. In this way, we make the network focus on the optimization of this occlusion sample. The results show that our method of hard sample mining is effective to improve the performance of our network.

\textbf{Data Augmentation:} Data augmentation can effectively prevent overfitting. To solve the limitation of training data, we adopt some data augmentation strategies, such as Random Erasing~\cite{zhong2020random}, Local Grayscale Transformation~\cite{Gong2021AGM}.

Random Erasing~\cite{zhong2020random}: In person ReID, persons in the images are sometimes occluded by other objects. To address the occlusion problem and improve the generalization ability of ReID models, Zhong et al. ~\cite{zhong2020random} proposed a new data augmentation approach named as Random Erasing Augmentation.

Local Grayscale Transformation(LGT)~\cite{Gong2021AGM}: Through the observation of the original dataset, we find that most of the personnel are players of the same team, so their dresses and appearances are very similar. Howerver, they have the different spatial structures. To address this problem, this method can be used as an effective data augmentation by introducing grayscale information, which proposed by Gong ~\cite{Gong2021AGM}. With this strategy, our model can achieve significant improvements in ReID task.

Besides the above two specific data augmentation methods, some regular augment methods are applied, such as random affine transformation, pixel padding, random flipping, etc.

\textbf{Imbalance Identities Resist:} Even if the data provider claims that in the training set, sequence is composed by 20 frames in the training set. But through statistics, we find that there are less than 20 images in IDs such as ID31 and ID164. We use the balanced ID data augmentation on the training set, copy few images up to 5 times, and several data augmentations are used for the copied images.

\begin{figure}[t]
\centering 
\includegraphics[width=0.45\textwidth]{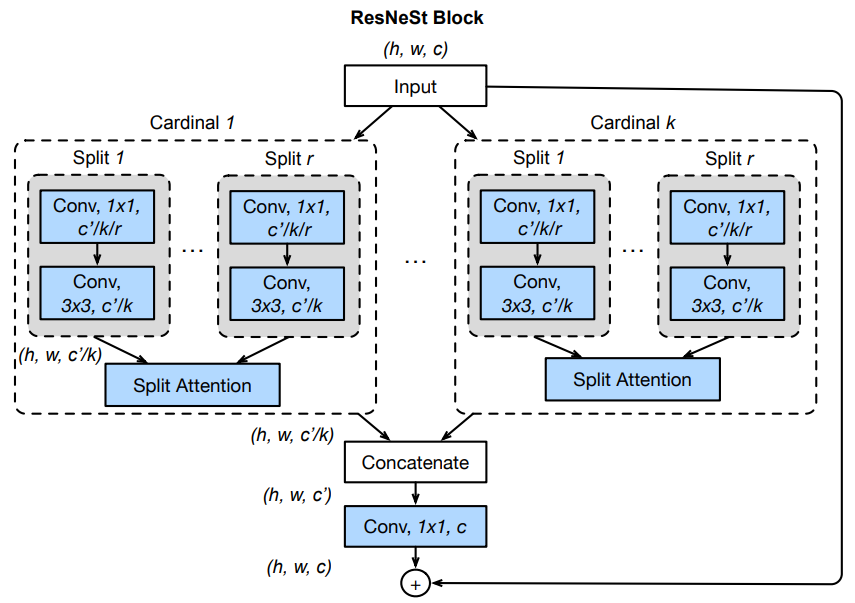}
\caption{Split-Attention Block of ResNeSt.}
\label{fig:fig3}
\end{figure}

\subsection{Model Architecture}
Baseline model is important for the final ranking, we use a CNN-based baseline ~\cite{luo2019bag} shown in Figure \ref{fig:fig2}. And besides ResNet~\cite{he2016deep}, we also use ResNeSt~\cite{zhang2020ResNeSt} and SE-ResNeXt~\cite{xie2017aggregated} as our backbones. These models are introduced as follows.

\textbf{ResNeSt:} The key part of ResNeSt is Split-Attention block. Split-Attention block is a computational unit consisting of feature-map group and split attention operations. Figure \ref{fig:fig3} depicts an overview of the Split-Attention Block.

\textbf{SE-ResNeXt:} SE-ResNeXt is constructed by repeating a building block that aggregates a set of transformations with the same topology. SE block~\cite{hu2018squeeze} adaptively recalibrates channel-wise feature responses by explicitly modelling interdependencies between channels. SE-ResNeXt has good performance in multiple tasks and is widely used.

\textbf{Other Techniques:} Various techniques are also applied. First, the aggregation layer aims to aggregate feature maps generated by the backbone into a global feature, and we use the  Generalized-Mean (GeM) pooling ~\cite{radenovic2018fine} instead of GAP proposed in ~\cite{luo2019bag}. Second, dropout is used for preventing overfitting. Finally, we remove the last spatial downsampling operation in the backbone network to increase the size of the feature map. For convenience, we denote the last spatial down-sampling operation in the backbone network as last stride.

\subsection{Loss function}
Categorical cross-entropy(CE) loss after softmax and triplet loss ~\cite{hermans2017defense} are widely used in person ReID task. But for VIPriors ReID dataset, we find that circle loss ~\cite{sun2020circle} performs better than CE loss. The total loss is proposed as follows for better performance. 
\begin{equation}
L = \alpha L_{Triplet} + \beta L_{circle}
\end{equation}

\textbf{Triplet Loss:} The triplet loss is computed as:
\begin{equation}
L_{Triplet} = [d_{p} - d_{n} + \alpha]_{+}
\end{equation}
where $d_{p}$ and $d_{n}$ are feature distances of positive pair and negative pair. $\alpha$ is the margin of triplet loss, and $[z]_{+}$ equals to $max(z, 0)$. In this paper, $\alpha$ is set to 0.4. However, triplet loss only considers the difference between $d_{p}$ and $d_{n}$ and ignores the absolute values of them.

\textbf{Circle Loss:} The circle loss benefits deep feature learning with high flexibility in optimization and more definite convergence target. It has a unified formula for two elemental learning approaches, i.e., learning with class-level labels and learning with pair-wise labels, more suitable for person ReID task. The derivation process of circle loss is not described here in detail, which can refer to ~\cite{sun2020circle}.

It is worth mentioning that in the Circle Loss, the relaxation factor $m$ is a very important hyper-parameter. We find that different $m$ may have different results in training, but this difference is very conducive to the ensemble of models. This point will be explained in detail in section 3.3.

\subsection{Post-processing}
Post-processing can significantly improve ReID performance in the inference stage. In this section, we will introduce several post-processing methods in this paper.

\textbf{Augmentation Test:} For each test image, we flip these two images horizontally and additionally extract two features. We get two features totally and then average them to obtain the final ReID feature.

\textbf{Re-Ranking:} We adopt a widely-used Re-ranking (RK) method ~\cite{zhong2017re} to update the final result. We notice that Jaccard distance is more suitable than Euclidean distance due to many people dressing in the same team uniform. We set $\lambda = 0.1$ in this paper.

\textbf{Query expansion:} We also use QE~\cite{chum2007total} to improve the performance of the retrieval system. But compared to re-ranking, the improvement of this technique is not significant.

\section{Experiments}
\subsection{Implementation Details}
All of our experiments are conducted in 8 NVIDIA A100 GPUs. In the training stage, we
train the model for 180 epochs with warming up ~\cite{goyal2017accurate} with initial learning rate of 0.0001 in first 10 epochs, increasing the learning rate from 0.0001 to 0.005 and dropout with probability of 0.2, weight decay of 0.0001 and label smooth are used for learning. In early stage, we trained model on training set for methods attempt and verification. And in the final stage, we trained models on both training set and validation set, and no external images or pre-trained weights are used. In addition, in order to obtain a better model, the model is trained several times by varying the random seed, and the results are combined together and treated as a whole, improving model mAP by 0.8\%.

\subsection{Results}
We show the ablation study of above different strategies in Table \ref{table:table1}. The baseline with ResNet50 trains on training set and evaluates on the validation set with the input size of $256 \times 128$. The baseline with ResNet50 backbone network can achieves 93.4708\% on validation set. Besides, it can be find that the strategies we proposed improve the performance by almost 18\% in terms of mAP, which shows the effectiveness of these techniques.

\begin{table}[h]
\setlength{\abovecaptionskip}{0.cm}
\setlength{\belowcaptionskip}{-0.9cm}
\caption{The ablation study of different strategies on validation set. With all methods, ResNet50 achieves 93.4708\% mAP.}
\label{table:table1}
\begin{tabular}{@{}c|c@{}}
\toprule
Method                                                 & mAP(\%) \\ \midrule
Baseline(ResNet50)                                     & 76.2044 \\
+ Noise Resist                                         & 78.0190 \\
+ Hard Sample Mining                                   & 80.9477 \\
+ Data Augmentation(Random Erasing and LGT)            & 81.4552 \\
+ Gem Pooling                                          & 83.8275 \\
+ Circle Loss                                          & 84.9631 \\
+ Augmentation Test                                    & 85.2631 \\
+ Re-rank                                              & 93.0423 \\
+ AQE                                                  & 93.1172 \\
+ Large Input Size($384\times128$)                     & 93.4708 \\ \bottomrule
\end{tabular}
\end{table}

\subsection{Ensembling}
Experimental evidence shows that the ensemble method is usually much more accurate than a single model. In our method, the ensemble method is the addition of the distance matrix of all prediction. For a better performance, we have ensembled predictions of above methods in total 24 models including ResNet101, ResNet152, ResNet200, ResNeSt-101, ResNeSt-152, ResNeSt-200, SE-ResNeXt101, SE-ResNeXt152, SE-ResNeXt200, with different input sizes and different relaxation factor $m$ of circle loss. 

\textbf{Tips:} We trained all models on both training set and validation set, for each backbone, we use two different sizes ($384\times128$ and $384\times192$) as input. Besides, different relaxation factor $m$ of circle loss will get different results, and we find that fusing the results has a certain improvement. Specifically, we set $m=0.3$ and $m=0.4$. Therefore, we fuse all the results as the final submission. The 96.5154\% in mAP is the final ensemble results, ranking first in the learderboard.

\section{Conclusion}
In our method, three strong network architectures were taken as the backbones. The usage of multiple data pre-processing and post-processing strategies improves the performance of the models. Besides, multiple testing methods and ensemble strategies improve the generalization and robustness of the models and prevent overfitting. Finally, we win the 1st place in VIPriors re-identification competition.

{\small
\bibliographystyle{ieee}
\bibliography{egbib}

\begin{thebibliography}{10}\itemsep=-1pt

\bibitem{chum2007total}
O.~Chum, J.~Philbin, J.~Sivic, M.~Isard, and A.~Zisserman.
\newblock Total recall: Automatic query expansion with a generative feature
  model for object retrieval.
\newblock In {\em 2007 IEEE 11th International Conference on Computer Vision},
  pages 1--8. IEEE, 2007.

\bibitem{Gong2021AGM}
Y.~Gong.
\newblock A general multi-modal data learning method for person
  re-identification.
\newblock 2021.

\bibitem{goyal2017accurate}
P.~Goyal, P.~Doll{\'a}r, R.~Girshick, P.~Noordhuis, L.~Wesolowski, A.~Kyrola,
  A.~Tulloch, Y.~Jia, and K.~He.
\newblock Accurate, large minibatch sgd: Training imagenet in 1 hour.
\newblock {\em arXiv preprint arXiv:1706.02677}, 2017.

\bibitem{he2016deep}
K.~He, X.~Zhang, S.~Ren, and J.~Sun.
\newblock Deep residual learning for image recognition.
\newblock In {\em Proceedings of the IEEE conference on computer vision and
  pattern recognition}, pages 770--778, 2016.

\bibitem{hermans2017defense}
A.~Hermans, L.~Beyer, and B.~Leibe.
\newblock In defense of the triplet loss for person re-identification.
\newblock {\em arXiv preprint arXiv:1703.07737}, 2017.

\bibitem{hu2018squeeze}
J.~Hu, L.~Shen, and G.~Sun.
\newblock Squeeze-and-excitation networks.
\newblock In {\em Proceedings of the IEEE conference on computer vision and
  pattern recognition}, pages 7132--7141, 2018.

\bibitem{liu2021hlfnet}
C.~Liu, L.-J. Guo, and R.~Zhang.
\newblock Hlfnet: High-low frequency network for person re-identification.
\newblock {\em IEEE Signal Processing Letters}, 2021.

\bibitem{luo2019bag}
H.~Luo, Y.~Gu, X.~Liao, S.~Lai, and W.~Jiang.
\newblock Bag of tricks and a strong baseline for deep person
  re-identification.
\newblock In {\em Proceedings of the IEEE/CVF Conference on Computer Vision and
  Pattern Recognition Workshops}, pages 0--0, 2019.

\bibitem{radenovic2018fine}
F.~Radenovi{\'c}, G.~Tolias, and O.~Chum.
\newblock Fine-tuning cnn image retrieval with no human annotation.
\newblock {\em IEEE transactions on pattern analysis and machine intelligence},
  41(7):1655--1668, 2018.

\bibitem{shrivastava2016training}
A.~Shrivastava, A.~Gupta, and R.~Girshick.
\newblock Training region-based object detectors with online hard example
  mining.
\newblock In {\em Proceedings of the IEEE conference on computer vision and
  pattern recognition}, pages 761--769, 2016.

\bibitem{sun2020circle}
Y.~Sun, C.~Cheng, Y.~Zhang, C.~Zhang, L.~Zheng, Z.~Wang, and Y.~Wei.
\newblock Circle loss: A unified perspective of pair similarity optimization.
\newblock In {\em Proceedings of the IEEE/CVF Conference on Computer Vision and
  Pattern Recognition}, pages 6398--6407, 2020.

\bibitem{sun2018beyond}
Y.~Sun, L.~Zheng, Y.~Yang, Q.~Tian, and S.~Wang.
\newblock Beyond part models: Person retrieval with refined part pooling (and a
  strong convolutional baseline).
\newblock In {\em Proceedings of the European conference on computer vision
  (ECCV)}, pages 480--496, 2018.

\bibitem{xie2017aggregated}
S.~Xie, R.~Girshick, P.~Doll{\'a}r, Z.~Tu, and K.~He.
\newblock Aggregated residual transformations for deep neural networks.
\newblock In {\em Proceedings of the IEEE conference on computer vision and
  pattern recognition}, pages 1492--1500, 2017.

\bibitem{zhang2020ResNeSt}
H.~Zhang, C.~Wu, Z.~Zhang, Y.~Zhu, H.~Lin, Z.~Zhang, Y.~Sun, T.~He, J.~Mueller,
  R.~Manmatha, et~al.
\newblock Resnest: Split-attention networks.
\newblock {\em arXiv preprint arXiv:2004.08955}, 2020.

\bibitem{zhao2017deeply}
L.~Zhao, X.~Li, Y.~Zhuang, and J.~Wang.
\newblock Deeply-learned part-aligned representations for person
  re-identification.
\newblock In {\em Proceedings of the IEEE international conference on computer
  vision}, pages 3219--3228, 2017.

\bibitem{zheng2015scalable}
L.~Zheng, L.~Shen, L.~Tian, S.~Wang, J.~Wang, and Q.~Tian.
\newblock Scalable person re-identification: A benchmark.
\newblock In {\em Proceedings of the IEEE international conference on computer
  vision}, pages 1116--1124, 2015.

\bibitem{zhong2017re}
Z.~Zhong, L.~Zheng, D.~Cao, and S.~Li.
\newblock Re-ranking person re-identification with k-reciprocal encoding.
\newblock In {\em Proceedings of the IEEE conference on computer vision and
  pattern recognition}, pages 1318--1327, 2017.

\bibitem{zhong2020random}
Z.~Zhong, L.~Zheng, G.~Kang, S.~Li, and Y.~Yang.
\newblock Random erasing data augmentation.
\newblock In {\em Proceedings of the AAAI Conference on Artificial
  Intelligence}, volume~34, pages 13001--13008, 2020.

\end{thebibliography}
}

\end{document}